\documentclass[sigconf,screen,nonacm]{acmart}
\usepackage{algorithm}
\usepackage{algpseudocode}
\usepackage{graphicx}
\usepackage{multirow}
\usepackage{eso-pic}  
\AtBeginDocument{%
  }

\begin{document}

\title{ AndroidControl-Curated: Revealing the True Potential of GUI Agents through Benchmark Purification}


\author{
    LEUNG Ho Fai (Kevin)$^{\dagger, *}$,
    XI XiaoYan (Sibyl)$^{\dagger}$,
    ZUO Fei (Eric)$^{\dagger}$
}

\affiliation{%
    \vspace{1em}
    \institution{$^{\dagger}$\LARGE BMW ArcherMind Information Technology Co. Ltd. (BA TechWorks)} 
    \country{} 
}

{\large\email{{kevin.leung, sibyl.xi, eric.zuo}@batechworks.com}}

\vspace{1em}
\thanks{$^*$Corresponding author. All authors contributed equally to this research.}

\renewcommand{\shortauthors}{Leung, et al.}

\begin{abstract}
On-device virtual assistants like Siri and Google Assistant are increasingly pivotal, yet their capabilities are hamstrung by a reliance on rigid, developer-dependent APIs. GUI agents offer a powerful, API-independent alternative, but their adoption is hindered by the perception of poor performance, as even the best models (e.g. Qwen3-VL-235B) scores are capped at around 60\% on benchmarks like AndroidControl, far from viability for real-world use.

Our research reveals that issue lies not only with the models but with the benchmarks themselves. We identified notable shortcomings in AndroidControl, including ambiguities and factual errors, which systematically underrates agent capabilities. To address this critical oversight, we enhanced AndroidControl into \textbf{AndroidControl-Curated}, a refined version of the benchmark improved through a rigorous purification pipeline. On this enhanced benchmark, state-of-the-art models achieve success rates nearing 75\% on complex tasks (15\% improvement), reflecting that on-device GUI agents are actually closer to practical deployment than previously thought. We introduce our new SOTA model, \textbf{Magma-R1-3B},  post-trained on just 2.4k curated samples using 60 hours of an H20 GPU (approximately \$60).  Despite being 200 times smaller in parameters, this model delivers performance comparable to Qwen3-VL-235B.  We release both \textbf{AndroidControl-Curated} benchmark and \textbf{Magma-R1} model to the research community, encouraging adoption of this enhanced benchmark to better reflect model capabilities and accelerate the development of robust, on-device virtual assistants.
\end{abstract}

%




\AddToShipoutPictureBG*{%
  \AtPageUpperLeft{%
    \put(\LenToUnit{0.3cm}, \LenToUnit{-1.8cm}){%
      \includegraphics[width=6cm]{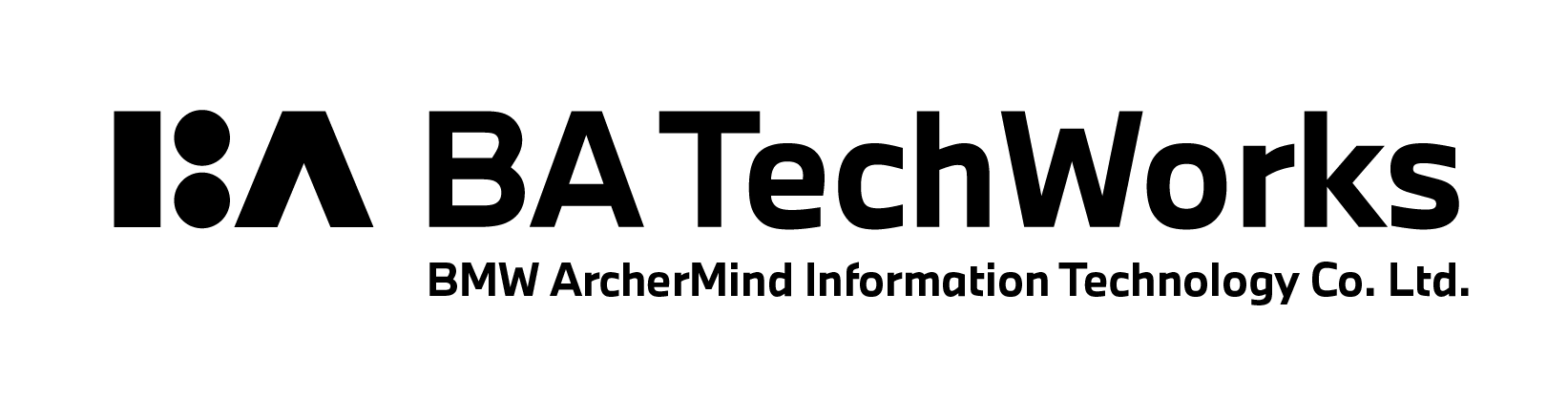}%
    }%
  }%
}
\maketitle

\noindent\textbf{Code and Data:} \url{https://github.com/batechworks/AndroidControl_Curated}

\begin{figure}[ht]
\centering
\includegraphics[width=\columnwidth]{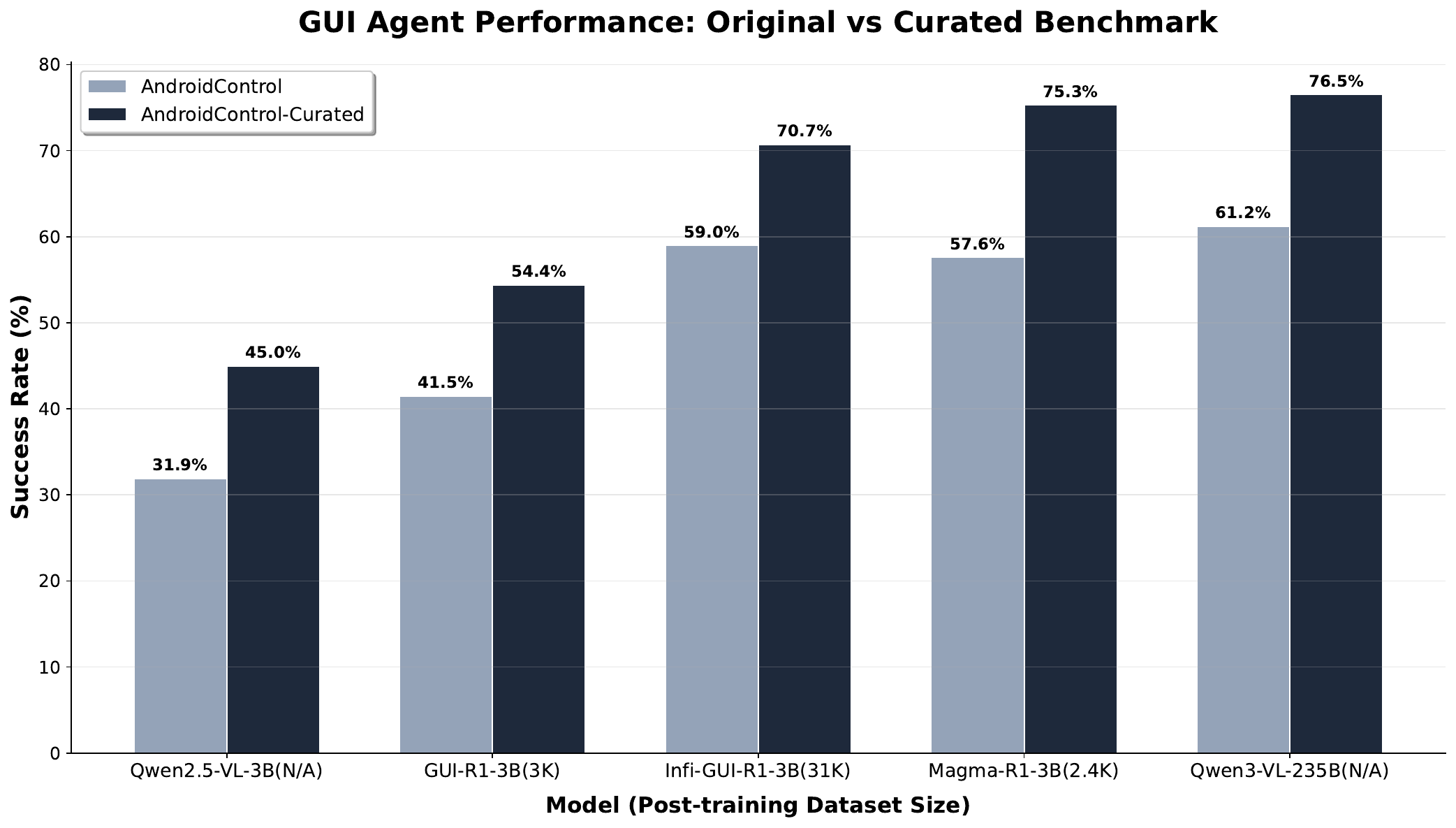}
\caption{\textbf{Benchmark Quality Reveals True Model Capabilities.} Performance comparison on AndroidControl (original) vs AndroidControl-Curated (purified). Our benchmark purification dramatically improves evaluation accuracy, revealing that compact 3B models (Magma-R1: 75.3\%) achieve performance remarkably close to large 235B models (Qwen3-VL-235B: 76.5\%) on challenging GUI tasks, up from an upper limit of around 60\% of even the world best models. This demonstrates that previous low performance scores were largely due to benchmark deficiencies rather than fundamental model limitations.}
\Description{A grouped bar chart comparing GUI agent performance on original AndroidControl benchmark versus AndroidControl-Curated, showing significant performance improvements across all models and revealing the true capabilities of compact models.}
\label{fig:benchmark_comparison}
\end{figure}

\begin{figure*}[ht]
\centering
\includegraphics[width=\textwidth]{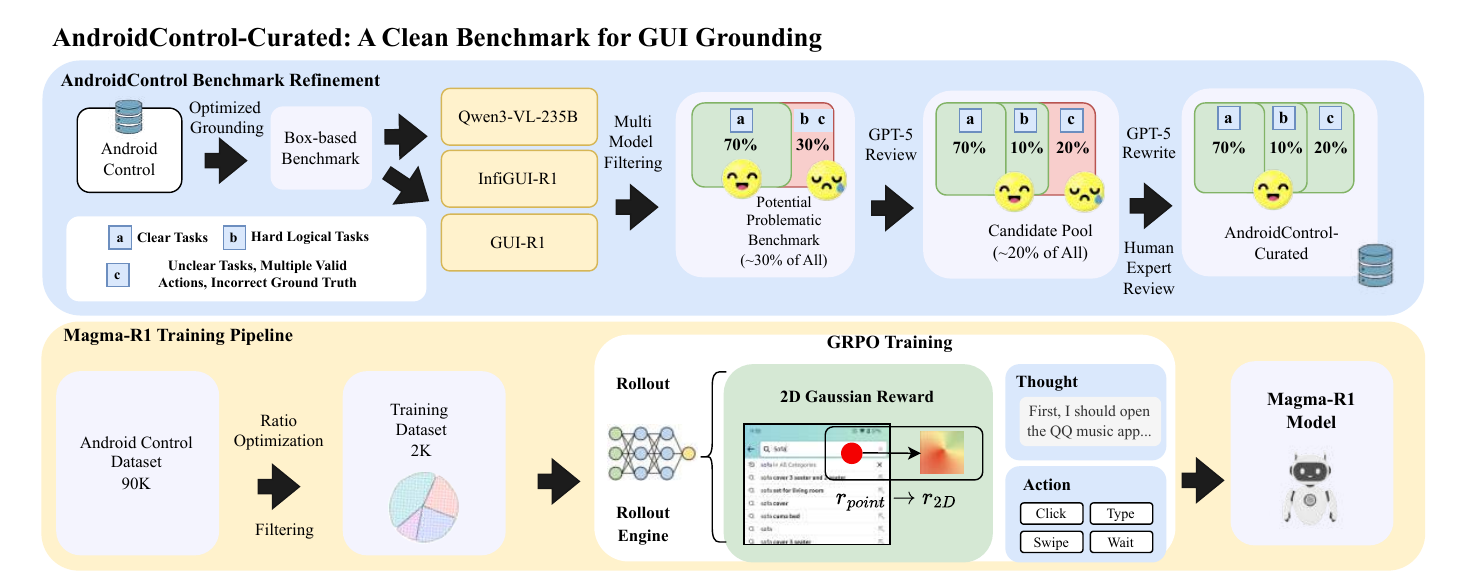}
\caption{Overview of our integrated pipeline. The upper part shows the Magma-R1 training process with GRPO training, featuring 2D Gaussian reward optimization and ratio optimization for balanced learning. The lower part illustrates the AndroidControl benchmark refinement process, including grounding optimization, multi-model filtering, LLM review and rewrite, leading to the creation of AndroidControl-Curated.}
\Description{A comprehensive system diagram showing two main components: Magma-R1 training pipeline with GRPO training and AndroidControl benchmark refinement process.}
\label{fig:overview}
\end{figure*}

\section{Introduction}
Virtual assistants like "Google Assistant" and "Siri" have become a key battleground for mobile phone and automotive manufacturers, increasingly central to the user experience and brand differentiation. Their value is particularly pronounced in cars, where hands-free voice commands for navigation, music, or vehicle functions are critical for safety. This has spurred a shift toward on-device models, such as Google's Gemini Nano (1.8–3.25B parameters) and Apple's 3B parameter foundation model, which are crucial for delivering the fast, private, and offline-capable interactions that users demand.

However, the capability of these on-device assistants remains severely limited. Due to constrained computational resources, they are restricted to simple tasks like setting timers or launching apps. For more complex, in-app actions, they rely on structured APIs like Apple's App Intents or Google's App Actions. This approach has a fundamental flaw: it is rigid and entirely dependent on developers explicitly providing support, leaving a vast ecosystem of legacy or unsupported applications beyond the assistant's reach.

GUI agents present a far more flexible and powerful alternative~\cite{tang_survey_2025}. By directly interacting with an application's user interface—reading screen content and programmatically performing actions like tapping and typing—they can automate tasks in any app, regardless of developer support~\cite{wang_mobile-agent_2024, cheng_seeclick_2024}. This paradigm excels at handling the dynamic, multi-step workflows inherent in modern applications, such as filling forms or navigating complex menus, without needing predefined API endpoints that often break with UI updates~\cite{zhang_you_2024, he_webvoyager_2024}.

Despite this potential, the prevailing view is that on-device GUI agents are not yet viable. Foundation models of a comparable size (3B parameters) have demonstrated low accuracy on high-level tasks, often hovering in the 60\% range on benchmarks like AndroidControl even after fine-tuning~\cite{li_effects_2024}.  Even the latest Qwen3-VL-235B, with 200 times the parameter, seems to reach the same ceiling at around 60\% success rate.  

This performance is far too low for a reliable user-facing product. This research was born from a direct confrontation with this problem. While developing post-training methodologies for compact vision-language models~\cite{wu_mobilevlm_2024, nong_mobileflow_2024}, we encountered perplexing phenomena: training with larger, seemingly higher-quality datasets from the benchmark sometimes led to \textit{reduced} accuracy.

This prompted a deep root-cause analysis, which uncovered a startling truth: the problem lay not with the models, but with the benchmark itself. We found that approximately 30\% of the AndroidControl benchmark, though derived from real-world human interactions, was riddled with ambiguities, multiple plausible solutions unaccounted for, and factual inaccuracies. These flaws were systematically penalizing correct and intelligent model behaviors, leading to a significant underestimation of their true capabilities.

To rectify this, we developed a rigorous, semi-automated benchmark purification pipeline, culminating in our primary contribution: \textbf{AndroidControl-Curated}. As demonstrated in Figure~\ref{fig:benchmark_comparison}, our purified benchmark provides a more realistic and reliable measure of an agent's abilities. The results are transformative: on \textbf{AndroidControl-Curated}, state-of-the-art model Infi-GUI-R1-3B achieves success rate exceeding 70\% (11\% increase) on challenging tasks, and the latest Qwen3-VL-235B model even achieves 76.5\% success rate (15\% increase).  This reveals that on-device GUI agents are actually far more capable than previously believed, making them a much nearer-term reality than suggested by earlier benchmarks.  This work brings us significantly closer to developing the next generation of GUI-agent-based virtual assistants.

Our main contributions are summarized as follows:
\begin{itemize}
    \item We identify and systematically analyze critical flaws in a mainstream GUI agent benchmark, revealing that poor benchmark quality is a key bottleneck hindering the perceived viability of on-device GUI agents.
    \item We propose a reproducible, semi-automated benchmark purification pipeline to address these flaws, providing a methodology for creating more reliable evaluation tools.
    \item We release \textbf{AndroidControl-Curated}, a rigorously purified benchmark. On this benchmark, we show that existing models are far more capable than previously believed, achieving performance that makes on-device GUI agents a practical possibility in the nearer future.
    \item We open-source \textbf{Magma-R1}, a new state-of-the-art model that achieves comparable results to models trained with 13X more data, as shown in Figure~\ref{fig:benchmark_comparison}, demonstrating that benchmark quality is more critical than model scale for GUI agent evaluation.
\end{itemize}

\begin{figure}[ht]
\centering
\includegraphics[width=0.8\columnwidth]{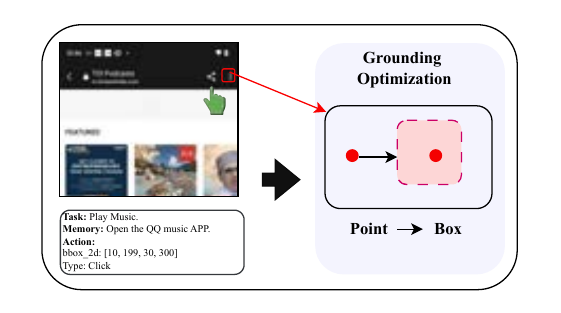}
\caption{Illustration of our grounding optimization approach. We transform the evaluation from exact point matching to bounding box-based intent alignment. The example shows a GUI task where the agent needs to interact with a UI element, and our method evaluates whether the predicted point falls within the target element's bounding box rather than matching exact coordinates.}
\Description{A diagram showing the transformation from point-based to box-based grounding evaluation, with a mobile interface example demonstrating the concept.}
\label{fig:grounding_optimization}
\end{figure}

\section{Methodology}

Our methodology consists of two main parts. First, we describe the systematic pipeline for constructing our high-quality benchmark, AndroidControl-Curated, designed to more accurately evaluate GUI agents. Second, we detail the training paradigm for our SOTA agent, Magma-R1, which leverages this curated data with our novel reinforcement learning strategies. The entire integrated process is summarized in Algorithm 1 and illustrated in Figure~\ref{fig:overview}.

\subsection{Systematic Benchmark Purification: The AndroidControl-Curated Pipeline}

While existing GUI agent benchmarks like AndroidControl have been invaluable resources for the field~\cite{li_effects_2024}, we find their evaluation methods and data labels contain systemic flaws. These deficiencies lead to a significant underestimation of agent capabilities and may misdirect future research~\cite{xu_mobile-bench-v2_2025, chai_a3_2025}. To resolve this, we propose a systematic, semi-automated benchmark purification pipeline to create a fairer, more accurate, and more challenging benchmark—**AndroidControl-Curated**. This process involves two core stages: grounding metric refinement and task-level correction.

\subsubsection{Stage 1: From Coordinate Matching to Intent Alignment in Grounding Evaluation}

\textbf{Problem Statement:} Original benchmarks typically evaluate grounding accuracy using \textbf{Exact Point Matching}~\cite{cheng_seeclick_2024, lin_showui_2024}. This approach is fundamentally flawed as it assesses the geometric proximity of predicted and labeled coordinates, not whether the model's \textit{interaction intent} is correct.

Formally, for a given GUI screenshot $I$ and a ground-truth interaction point $p_{gt} = (x_{gt}, y_{gt})$, the model's prediction is $p_{pred}$. The traditional evaluation function, $\mathcal{E}_{\text{point}}$, can be defined as:
\[
\mathcal{E}_{\text{point}}(p_{pred}, p_{gt}) = \mathbb{I}(\| p_{pred} - p_{gt} \|_2 \le \tau)
\tag{4}
\]
where $\mathbb{I}(\cdot)$ is the indicator function and $\tau$ is a minimal tolerance threshold, often set to 0. This metric is overly strict, as it ignores the fact that any click within the area of an interactive UI element should be considered a valid execution of a human's intent. Therefore, $\mathcal{E}_{\text{point}}$ fails to accurately measure the model's understanding of the correct interaction target.

\textbf{Solution: Bounding-Box-based Intent Alignment.} To align the evaluation metric more closely with human interaction logic, we propose a new paradigm based on UI element bounding boxes, as illustrated in Figure~\ref{fig:grounding_optimization}~\cite{gou_navigating_2025, yang_aria-ui_2024}. The core idea is to shift the focus of evaluation from the geometric alignment of coordinates to the semantic alignment of intent.

Specifically, we leverage the Document Object Model (DOM) or Accessibility Tree of the GUI~\cite{huang_spiritsight_2025, nayak_ui-vision_nodate}. For each interactive UI element $e_k$ on screen $I$, we can obtain its minimal bounding box $B_k$. We define a mapping function $\mathcal{F}: p \rightarrow B$ that maps any point $p$ to the minimal UI element's bounding box $B$ that contains it.

Consequently, we upgrade the ground-truth point $p_{gt}$ to a more robust \textbf{Ground-Truth Interaction Area}, $B_{gt} = \mathcal{F}(p_{gt})$. Our new grounding evaluation function, $\mathcal{E}_{\text{bbox}}$, is defined as:
\[
\mathcal{E}_{\text{bbox}}(p_{pred}, p_{gt}) = \mathbb{I}(p_{pred} \in B_{gt}) = \mathbb{I}(p_{pred} \in \mathcal{F}(p_{gt}))
\tag{5}
\]
This function checks if the predicted point $p_{pred}$ falls within the bounding box $B_{gt}$ of the UI element corresponding to the ground-truth point $p_{gt}$. This refinement offers significant advantages in robustness, intent alignment, and fairness. We refer to the benchmark version with only this refinement applied as \textbf{AndroidControl-Curated-Box}.

\subsubsection{Stage 2: Task-Level Correction via LLM-Human Collaboration}

After addressing the evaluation-level bias, we tackle a deeper challenge: \textbf{factual deficiencies} within the benchmark data itself~\cite{liu_learnact_2025, zhang_tongui_2025}. To eradicate these issues, we designed a three-step, semi-automated task-level correction process.

\paragraph{Step 1: High-Risk Sample Identification via Execution Consensus Failure} To efficiently locate potentially flawed samples, we employ a heuristic strategy based on \textbf{Execution Consensus Failure}~\cite{yang_zerogui_2025, fan_gui-bee_2025}. The assumption is that if a task cannot be solved by multiple, diverse, high-performing agents—in our case, Qwen3-VL-235B, InfiGUI-R1, and GUI-R1—the task itself is more likely to be flawed than all agents being incapable.

Formally, we define a set of $N$ top-tier GUI agents $\mathcal{M} = \{M_1, \dots, M_N\}$. For any task $T_j$ in the AndroidControl $\mathcal{D}_{\text{orig}}$, we define a success function $S(M, T) \in \{0, 1\}$. The high-risk candidate set $\mathcal{D}_{\text{cand}}$ is then identified as:
\[
\mathcal{D}_{\text{cand}} = \left\{ T_j \in \mathcal{D}_{\text{orig}} \mid \sum_{k=1}^{N} S(M_k, T_j) = 0 \right\}
\tag{6}
\]
This strategy effectively pinpoints tasks that were failed by every agent in our expert set, significantly narrowing the scope for manual review.

\paragraph{Step 2: Automated Causal Attribution and Correction with LLMs} We leverage a large language model (LLM) as an \textbf{Automated Reviewer}, $\mathcal{L}_{\text{reviewer}}$~\cite{liu_ui-e2i-synth_nodate, jiang_appagentx_2025}. After establishing a comprehensive \textbf{Deficiency Taxonomy} $\mathcal{C}$ (e.g., Multiple Valid Actions, Unclear Task, Wrong Ground Truth), we feed each candidate task $T_j$ along with its context and the failed execution traces from our agent set $\mathcal{M}$ to the reviewer. This process is a mapping that yields a structured analysis tuple $A_j$:
\[
A_j = (c_j, \Delta t_j, \Delta g_j, \rho_j)
\tag{7}
\]
where $c_j \in \mathcal{C}$ is the causal attribution of the failure, $\Delta t_j$ is a proposed revision for the task instruction, $\Delta g_j$ is a proposed revision for the ground-truth trajectory, and $\rho_j$ is a natural language rationale. This allows for in-depth, scalable analysis and correction proposals.

\paragraph{Step 3: Rigorous Human Expert Verification} Finally, all correction proposals generated by $\mathcal{L}_{\text{reviewer}}$ are subject to a \textbf{Human-in-the-Loop} verification process~\cite{wang_mobile-agent-e_2025}. A panel of human experts reviews a large, random sample of the automated corrections, validating both the causal attribution and the quality of the proposed fixes. This final step ensures the reliability and quality of the resulting benchmark, \textbf{AndroidControl-Curated}.

\subsection{Training Paradigm of Magma-R1: Optimization via GRPO}

To enhance the decision-making and operational precision of GUI agents in complex tasks, we employ a policy optimization-based reinforcement learning framework to train Magma-R1~\cite{lu_ui-r1_2025, luo_gui-r1_2025}. We opt for an efficient policy gradient algorithm, which we term Generative REINFORCE with Policy Optimization (GRPO). Its core idea is inspired by Proximal Policy Optimization (PPO), ensuring training stability by constraining the policy update step size~\cite{yuan_enhancing_2025}.

The objective function of GRPO is defined as:
\begin{align}
\mathcal{J}_{\text{GRPO}}(\theta) &= \mathbb{E}_{q \sim \mathcal{D}, \{o_i\}_{i=1}^G \sim \pi_{\theta_{\text{old}}}(\cdot|q)} \left[ \frac{1}{G} \sum_{i=1}^{G} \right. \nonumber \\
&\quad \left. \left( \min \left( r_i(\theta) A_i, \text{clip}(r_i(\theta), 1-\epsilon, 1+\epsilon) A_i \right) \right) \right]
\tag{1}
\end{align}
where $r_i(\theta) = \frac{\pi_{\theta}(o_i|q)}{\pi_{\theta_{\text{old}}}(o_i|q)}$ is the probability ratio for action $o_i$ between the new and old policies, $\pi_{\theta}$ and $\pi_{\theta_{\text{old}}}$. $A_i$ is the advantage function, which evaluates the quality of action $o_i$ relative to the average for a given query $q$. $\epsilon$ is a hyperparameter that defines the clipping range to prevent excessively large policy updates.

In our implementation, the advantage function $A_i$ is estimated by normalizing the in-batch rewards as follows:
\[
A_i = \frac{r_i - \text{mean}(\{r_1, r_2, ..., r_G\})}{\text{std}(\{r_1, r_2, ..., r_G\}) + \delta}
\tag{2}
\]
where $r_i$ is the reward for action $o_i$, $G$ is the batch size, and $\delta$ is a small constant for numerical stability. This method uses the average reward of the batch as a baseline, effectively reducing the variance of the policy gradient.

However, the traditional reinforcement learning paradigm faces two significant challenges in GUI tasks: **Reward Sparsity** and **Data Imbalance**. To address these, we introduce two key optimizations.

\subsubsection{Dense Rewards: Gaussian Kernel-based Grounding Reward}

In GUI grounding tasks, conventional reward functions (e.g., based on IoU or whether a point is within a bounding box) are typically binary (0 or 1), leading to a severe reward sparsity problem~\cite{cheng_seeclick_2024, gou_navigating_2025}. The model receives positive feedback only for predictions within the exact correct region, while "nearly correct" predictions are treated as complete failures, providing no effective learning signal.

To solve this, we design a smooth reward function based on a 2D Gaussian kernel, which provides a dense and continuous reward signal for grounding tasks~\cite{luo_visual_2025, tang_think_2025}. Specifically, for a task requiring the localization of a target point $p_{gt} = (x_{gt}, y_{gt})$, the model predicts a bounding box $B_{pred}$. We first calculate the center of this box, $c_{pred} = (c_x, c_y)$. The reward function $R_{\text{grounding}}$ is defined as:
\[
R_{\text{grounding}}(B_{pred}, p_{gt}) = \exp \left( - \frac{\| c_{pred} - p_{gt} \|_2^2}{2\sigma^2} \right)
\tag{3}
\]
where $\| \cdot \|_2^2$ denotes the squared Euclidean distance, and $\sigma$ is a hyperparameter controlling the "tolerance" of the Gaussian function. This function has several advantages:
\begin{itemize}
    \item \textbf{Reward Density}: Any prediction receives a reward in the range (0, 1], decreasing with distance from the ground truth, thus solving the sparsity issue.
    \item \textbf{Smoothness}: The reward function is continuous and differentiable, providing a smooth optimization landscape for the policy gradient.
    \item \textbf{Physical Intuition}: It aligns with human interaction logic—a click slightly off-center but still on the target element is far better than a click that misses entirely.
\end{itemize}

\begin{algorithm*}[ht]
\caption{Our Integrated Pipeline: Agent Training and Benchmark Purification}
\label{alg:full_pipeline}
\begin{algorithmic}[1]
\State \textbf{Input:} AndroidControl $\mathcal{D}_{\text{orig}}$, un-trained agent parameters $\theta_0$
\State \textbf{Output:} Purified benchmark AndroidControl-Curated, trained agent Magma-R1
\Statex
\Function{CreateAndroidControlPro}{$\mathcal{D}_{\text{orig}}$, expert agents $\mathcal{M}$} \Comment{\textbf{Part 1: Benchmark Purification (Section 3.1)}}
    \State $\mathcal{D}_{\text{G}} \gets$ Apply bounding-box evaluation $\mathcal{E}_{\text{bbox}}$ to $\mathcal{D}_{\text{orig}}$ \Comment{Stage 1: Grounding refinement}
    \State $\mathcal{D}_{\text{cand}} \gets \{T_j \in \mathcal{D}_{\text{G}} \mid \forall M_k \in \mathcal{M}, S(M_k, T_j) = 0\}$ \Comment{Stage 2: Task correction}
    \State $\mathcal{C}_{\text{corrections}} \gets \emptyset$
    \For{each $T_j \in \mathcal{D}_{\text{cand}}$}
        \State $A_j \gets \mathcal{L}_{\text{reviewer}}(\text{context}(T_j), \text{failure traces})$
        \If{$A_j$ indicates data deficiency}
            \State $\mathcal{C}_{\text{corrections}} \gets \mathcal{C}_{\text{corrections}} \cup \{A_j\}$
        \EndIf
    \EndFor
    \State $\mathcal{C}_{\text{verified}} \gets \text{HumanExpertVerification}(\mathcal{C}_{\text{corrections}})$
    \State AndroidControl-Curated $\gets \text{ApplyCorrections}(\mathcal{D}_{\text{G}}, \mathcal{C}_{\text{verified}})$
    \State \Return AndroidControl-Curated
\EndFunction
\Statex
\Function{TrainMagmaR1}{$\mathcal{D}_{\text{orig}}, \theta_0$} \Comment{\textbf{Part 2: Agent Training with GRPO (Section 3.2)}}
    \State $\theta \gets \theta_0$
    \For{each training iteration}
        \State $\mathcal{B} \gets \text{StratifiedSample}(\mathcal{D}_{\text{orig}}, \mathcal{P}_{\text{target}})$ \Comment{Proportional sampling}
        \For{each query $q \in \mathcal{B}$}
            \State Generate outputs $\{o_i\}_{i=1}^G \sim \pi_{\theta_{\text{old}}}(\cdot|q)$
            \State Calculate dense rewards $\{r_i\}_{i=1}^G$ using Gaussian kernel (Eq. 3)
        \EndFor
        \State Compute advantages $\{A_i\}_{i=1}^G$ using Eq. 2
        \State Update $\theta$ by optimizing $\mathcal{J}_{\text{GRPO}}(\theta)$ (Eq. 1)
    \EndFor
    \State \Return trained agent Magma-R1
\EndFunction
\end{algorithmic}
\end{algorithm*}

\subsubsection{Balanced Learning: Action Type Proportional Optimization}

GUI operation datasets commonly suffer from a severe class imbalance problem, where "click/tap" actions constitute the vast majority, while other critical actions like "type", "scroll", and "swipe" are relatively rare~\cite{li_effects_2024, xu_mobile-bench-v2_2025}. This long-tailed distribution causes the model to over-focus on high-frequency actions during training, neglecting the learning of low-frequency yet crucial skills.

To promote balanced learning across all action types, we propose an \textbf{Action Type Proportional Optimization} strategy~\cite{wu_reachagent_2025, dai_advancing_2025}. This strategy employs stratified sampling based on a predefined \textbf{Target Distribution} when constructing training batches, rather than purely random sampling.

Formally, let $\mathcal{A} = \{a_1, ..., a_K\}$ be the set of all $K$ action types. In the original dataset $\mathcal{D}_{\text{orig}}$, the empirical distribution is $\mathcal{P}_{\text{orig}} = \{p(a_1), ..., p(a_K)\}$, where $p(a_{\text{click}})$ is much larger than other probabilities. Our goal is to define a more balanced target distribution $\mathcal{P}_{\text{target}}$, such that:
\[
\mathcal{P}_{\text{target}}(a_k) > \mathcal{P}_{\text{orig}}(a_k), \quad \forall a_k \in \{\text{type}, \text{scroll}, \dots\}
\]
In each training iteration, we sample from data pools stratified by action type, ensuring that the action distribution within each batch $\mathcal{B}$ approximates $\mathcal{P}_{\text{target}}$. This significantly increases the model's exposure to low-frequency actions, leading to a more comprehensive agent proficient in all GUI skills.

\section{Experiments}

In this section, we present a series of comprehensive experiments to validate the effectiveness of our methodology. Our experiments are designed to answer three core questions: (1) Does our proposed benchmark purification pipeline significantly improve the fairness and accuracy of evaluation? (2) What is the state-of-the-art performance of existing GUI agents on the purified new benchmark? (3) Can our Magma-R1 model, trained on a small amount of high-quality data, outperform baseline models trained on large-scale original data?

\begin{table*}[ht]
\centering
\caption{Performance comparison of GUI agents on AndroidControl-Curated. Grounding Accuracy (GA) for all models is evaluated using our proposed $\mathcal{E}_{\text{bbox}}$. The best results are in \textbf{bold}, and the second best are \underline{underlined}. "-" indicates results to be added.}
\label{tab:main_results}
\resizebox{\textwidth}{!}{%
\begin{tabular}{lcccccc}
\toprule
\multirow{2}{*}{\textbf{Model}} & \multicolumn{3}{c}{\textbf{AndroidControl-Curated-Easy}} & \multicolumn{3}{c}{\textbf{AndroidControl-Curated-Hard}} \\
\cmidrule(lr){2-4} \cmidrule(lr){5-7}
& Type (\%) & Grounding (\%) & SR (\%) & Type (\%) & Grounding (\%) & SR (\%) \\
\midrule
\multicolumn{7}{l}{\textit{Proprietary Models}} \\
GPT-4o & 74.3 & 0.0 & 19.4 & 66.3 & 0.0 & 20.8 \\
\midrule
\multicolumn{7}{l}{\textit{Open-source Models}} \\
OS-Atlas-4B & \textbf{91.9} & 83.8 & 80.6 & 84.7 & 73.8 & 67.5 \\
UI-R1 & 62.2 & 93.6 & 58.9 & 54.4 & 79.3 & 43.6 \\
GUI-R1-3B & 69.5 & \underline{94.7} & 67.1 & 63.1 & 80.3 & 54.4 \\
GUI-R1-7B & 74.9 & \textbf{95.9} & 72.7 & 66.5 & 82.6 & 57.5 \\
Infi-GUI-R1 (trained on 31k origin data) & 90.2 & 93.7 & \underline{87.2} & 78.5 & 72.8 & 70.7 \\
Qwen3-VL-30B & 82.8 & 80.7 & 70.5 & \underline{85.9} & 78.9 & 70.0 \\
Qwen3-VL-235B & 85.1 & 82.9 & 74.5 & \textbf{88.2} & \underline{83.6} & \textbf{76.5} \\
\midrule
\multicolumn{7}{l}{\textit{Ours}} \\
Magma-R1 & \underline{91.3} & 94.2 & \textbf{88.0} & 84.2 & \textbf{84.8} & \underline{75.3} \\
\bottomrule
\end{tabular}%
}
\end{table*}

\subsection{Experimental Setup}

\textbf{Benchmarks:} To systematically evaluate the impact of our work, our experiments revolve around a subset of the `AndroidControl` benchmark~\cite{li_effects_2024}, introducing two intermediate versions and one final version from our pipeline:
\begin{itemize}
    \item \textbf{AndroidControl:} The original, unmodified `AndroidControl Hard` benchmark. It serves as the starting point for all our comparisons to reveal the inherent flaws of existing benchmarks.
    \item \textbf{AndroidControl-Curated-Box:} This benchmark applies only the \textbf{grounding metric refinement} described in Section 3.2.1 to `AndroidControl`. The underlying data is identical to the original, but it is evaluated using our more robust $\mathcal{E}_{\text{bbox}}$ standard. This version aims to independently quantify the performance difference caused by the evaluation method itself.
    \item \textbf{AndroidControl-Curated:} This is our final, fully purified benchmark, having undergone the complete pipeline (both grounding metric refinement and task-level correction). It includes both `Easy` and `Hard` subsets from `AndroidControl`, providing a more accurate and reliable environment for evaluating GUI agents.
\end{itemize}

\textbf{Compared Models:} For a thorough comparison, we include several representative GUI agent models as baselines against our proposed model~\cite{liu_infigui-r1_2025, qin_ui-tars_2025}:
\begin{itemize}
    \item \textbf{Infi-GUI-R1:} A public, state-of-the-art model~\cite{liu_infigui-r1_2025} trained on over 31k original `AndroidControl` data points, serving as a key baseline to measure performance improvements.
    \item \textbf{Other Baselines:} We also include several other representative models such as UI-R1~\cite{lu_ui-r1_2025}, GUI-R1~\cite{luo_gui-r1_2025}, OS-Atlas, and UI-TARS~\cite{qin_ui-tars_2025} to ensure the breadth of our evaluation.
    \item \textbf{Magma-R1 (Ours):} Our proposed model, based on a 3B-parameter multimodal model~\cite{wu_mobilevlm_2024} and trained using the GRPO framework described in Section 3.1. Notably, this model was trained on \textbf{only 2,400 high-quality samples} selected from AndroidControl-Curated, designed to test the core hypothesis that "data quality trumps quantity."
\end{itemize}

\begin{table*}[ht]
\centering
\caption{Ablation analysis of the benchmark purification process on the Hard subset. SR Impr. (G) shows the SR gain from AndroidControl to AndroidControl-Curated-Box. SR Impr. (T) shows the SR gain from AndroidControl-Curated-Box to the final AndroidControl-Curated. Best results are in \textbf{bold}, second best are \underline{underlined}.}
\label{tab:ablation_study}
\resizebox{\textwidth}{!}{%
\begin{tabular}{l|ccc|cccc|cccc}
\toprule
\multirow{2}{*}{\textbf{Model}} & \multicolumn{3}{c|}{\textbf{AndroidControl}} & \multicolumn{4}{c|}{\textbf{AndroidControl-Curated-Box}} & \multicolumn{4}{c}{\textbf{AndroidControl-Curated}} \\
\cmidrule(lr){2-4} \cmidrule(lr){5-8} \cmidrule(lr){9-12}
& Type (\%) & Grounding (\%) & SR (\%) & Type (\%) & Grounding (\%) & SR (\%) & SR Impr. (G) & Type (\%) & Grounding (\%) & SR (\%) & SR Impr. (T) \\
\midrule
GUI-R1-3B & 57.2 & 59.0 & 41.5 & 59.3 & 74.0 & 49.4 & \textcolor{blue}{+7.9} & 63.1 & 80.3 & 54.4 & \textcolor{blue}{+5.0} \\
GUI-R1-7B & 62.5 & 65.1 & 46.3 & 63.3 & 76.9 & 53.2 & \textcolor{blue}{+6.9} & 66.5 & 82.6 & 57.5 & \textcolor{blue}{+4.3} \\
Infi-GUI-R1 & \underline{77.0} & 57.0 & \underline{59.0} & 77.7 & 69.5 & 67.6 & \textcolor{blue}{+8.6} & 78.5 & 72.8 & 70.7 & \textcolor{blue}{+3.1} \\
Qwen3-VL-235B & 67.3 & \textbf{78.3} & \textbf{61.2} & \textbf{82.9} & \textbf{79.9} & \textbf{71.7} & \textcolor{blue}{+10.5} & \textbf{88.2} & \underline{83.6} & \textbf{76.5} & \textcolor{blue}{+4.8} \\
\midrule
Magma-R1 & \textbf{78.2} & 58.2 & 57.6 & \underline{80.0} & \underline{77.1} & \underline{69.1} & \textcolor{blue}{+11.5} & \underline{84.2} & \textbf{84.8} & \underline{75.3} & \textcolor{blue}{+6.2} \\
\bottomrule
\end{tabular}%
}
\end{table*}

\textbf{Evaluation Metrics:} We use two complementary metrics for a comprehensive performance assessment, following established practices in GUI agent evaluation~\cite{cheng_seeclick_2024, wang_mobile-agent_2024}:
\begin{itemize}
    \item \textbf{Success Rate (SR):} The ultimate metric for evaluating an agent's ability to complete a task~\cite{zhang_you_2024}. A task is considered successful only if the agent executes all steps correctly and reaches the intended final screen state. This metric measures the agent's high-level, end-to-end planning and execution capabilities.
    \item \textbf{Grounding Accuracy (GA):} This metric measures the agent's ability to localize the target for an operation at each step~\cite{lin_showui_2024}. We consistently use the bounding-box-based intent alignment evaluation function, $\mathcal{E}_{\text{bbox}}$, defined in Section 3.2.1, to calculate GA for all models across all benchmarks. This ensures a fair and consistent evaluation of grounding ability throughout our experiments.
\end{itemize}

\subsection{Main Experiment: SOTA Performance on AndroidControl-Curated}

We first conduct a comprehensive performance evaluation of all models on the fully purified final benchmark, AndroidControl-Curated. This experiment aims to establish a new standard in two respects: first, to determine the new state-of-the-art performance on the fairest and most accurate GUI benchmark available; and second, to validate our advocated paradigm of "high-quality, small-sample training" through direct comparison with baselines trained on large-scale data~\cite{yuan_enhancing_2025}. The results are presented in Table~\ref{tab:main_results}.

\subsection{Ablation Study on Benchmark Purification}

To isolate and quantify the specific contribution of each stage in our benchmark purification pipeline, we conduct an ablation study~\cite{xu_mobile-bench-v2_2025}. In this experiment, we evaluate all models on three progressive versions of the `AndroidControl-Curated-Hard` benchmark: `AndroidControl`, `AndroidControl-Curated-Box`, and the final `AndroidControl-Curated`. `AndroidControl` uses the original, point-based matching for evaluation~\cite{cheng_seeclick_2024}, whereas `AndroidControl-Curated-Box` and `AndroidControl-Curated` both use our intent-aligned, bounding-box-based evaluation ($\mathcal{E}_{\text{bbox}}$). The results, presented in Table~\ref{tab:ablation_study}, clearly illustrate the evolution of model performance as the benchmark is progressively refined.

\subsection{Case Study: A Qualitative Analysis of Benchmark Deficiencies}

\begin{figure*}[ht]
\centering
\includegraphics[width=\textwidth]{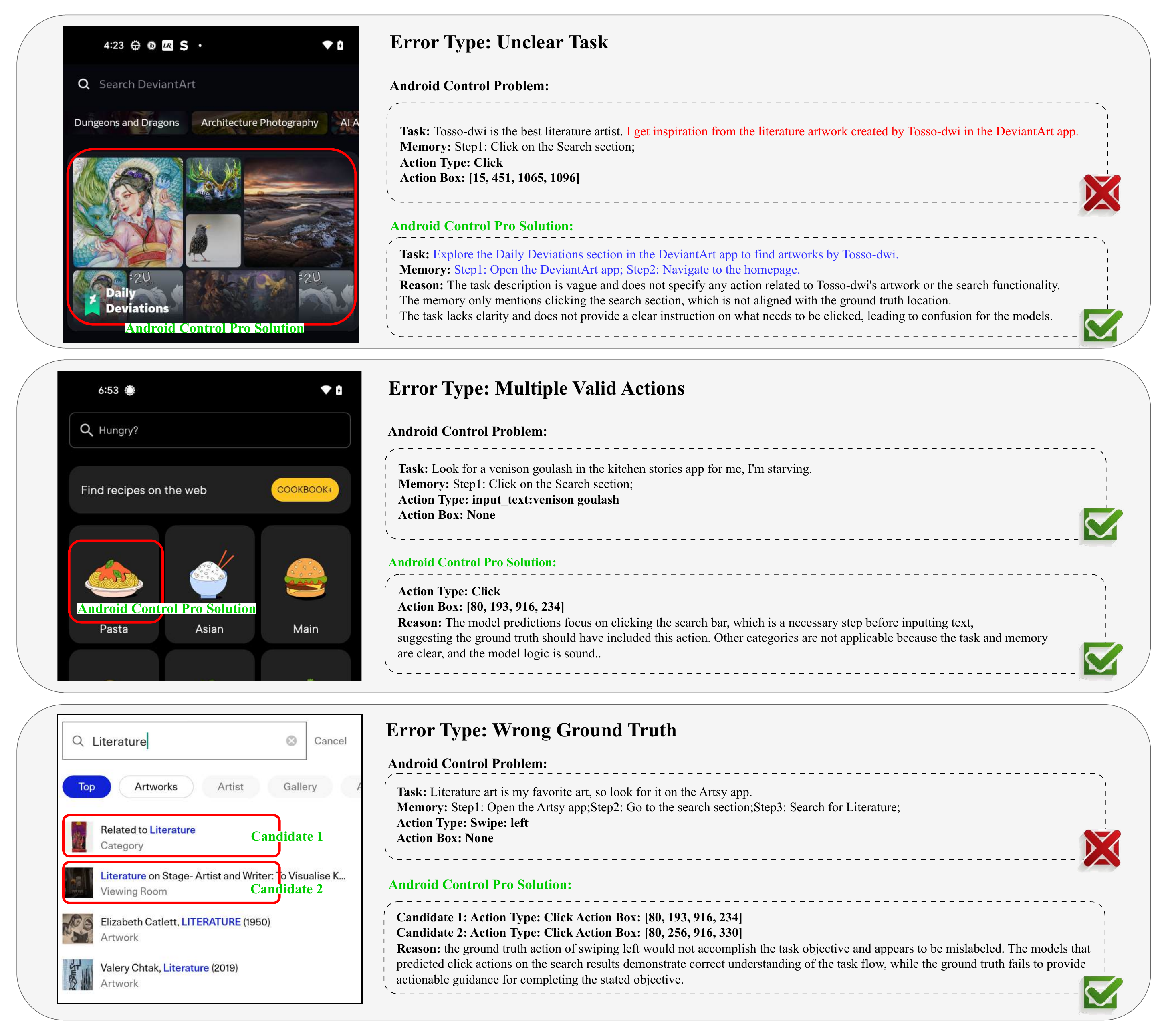}
\caption{\textbf{Case Study: Three Major Categories of Benchmark Deficiencies.} We show representative examples of the systematic errors found in the original AndroidControl benchmark: (Top) \textbf{Unclear Task} - vague task descriptions that lack clear navigation instructions; (Middle) \textbf{Multiple Valid Actions} - multiple valid actions for achieving the same goal; (Bottom) \textbf{Wrong Ground Truth} - incorrect labels that contradict the current UI state. Our AndroidControl-Curated corrections address each category by providing clearer instructions, accepting multiple valid actions, and fixing factual errors.}
\Description{A three-panel figure showing mobile app interfaces with annotations highlighting different types of benchmark deficiencies and their corrections in AndroidControl-Curated.}
\label{fig:case_study}
\end{figure*}

To visually demonstrate the necessity and effectiveness of our task-level correction process, this section provides a qualitative analysis of several typical error cases identified from `AndroidControl`~\cite{chai_a3_2025}. These cases were first flagged by our automated reviewer, $\mathcal{L}_{\text{reviewer}}$, and subsequently confirmed by human experts~\cite{wang_mobile-agent-e_2025}. Figure~\ref{fig:case_study} illustrates three representative categories of deficiencies we discovered and corrected.

\subsubsection{Case in Point: Three Categories of Systematic Deficiencies}

As illustrated in Figure~\ref{fig:case_study}, our analysis revealed three primary categories of deficiencies that collectively affect over 70\% of the AndroidControl samples, consistent with findings in other evaluation studies~\cite{liu_learnact_2025}.

\textbf{Unclear Task (24.13\% of samples):} The top panel shows a DeviantArt interface where the task instructs to "get inspiration from literature artwork created by Tosso-dwi," but provides no clear guidance on how to navigate to this specific content. This type of ambiguous instruction has been identified as a common issue in GUI benchmarks~\cite{zhang_tongui_2025}. The ground truth simply indicates clicking the search section, leaving models to guess the subsequent steps. Our correction provides explicit instructions: "Explore the Daily Deviations section in the DeviantArt app to find artworks by Tosso-dwi," giving models a clear, actionable path to success.

\textbf{Multiple Valid Actions (8.12\% of samples):} The middle panel demonstrates a cooking app scenario where the task is to "Look for a venison goulash recipe." While the ground truth specifies typing text, clicking on the search bar is an equally valid and necessary prerequisite action. This reflects the multi-modal nature of GUI interactions~\cite{yang_aria-ui_2024}. Our correction accepts both the search bar click and the subsequent text input as valid actions, recognizing that real user interactions often involve multiple valid pathways to achieve the same goal~\cite{huang_spiritsight_2025}.

\textbf{Wrong Ground Truth (37.63\% of samples):} The bottom panel shows the most prevalent issue—factually incorrect labels, a problem that has been noted in other benchmark studies~\cite{liu_ui-e2i-synth_nodate}. In an Artsy app interface with literature search results already displayed, the ground truth inexplicably specifies "Swipe: left" with no bounding box, an action that would not advance the stated goal of finding literature art. Our correction identifies that clicking on the displayed search results (such as "Related to Literature") represents the logical next step, replacing the nonsensical ground truth with actionable candidates.

The systematic nature of these deficiencies explains why training on larger datasets sometimes reduced model performance—the benchmark was actively teaching incorrect behaviors~\cite{dai_advancing_2025}. By correcting these issues in AndroidControl-Curated, we observe significant improvements: tasks previously marked as 0\% successful due to rigid evaluation criteria now achieve success rates approaching 75\%, revealing the true capabilities of existing models and enabling more effective training of new ones~\cite{wu_reachagent_2025}.

This analysis demonstrates that benchmark quality, not model architecture limitations, was the primary bottleneck preventing the practical deployment of on-device GUI agents~\cite{fan_gui-bee_2025}. Our systematic correction methodology provides a replicable framework for improving evaluation reliability across the broader field~\cite{yang_zerogui_2025}.

\section{Related Work}

\subsection{Multimodal Large Language Models}

Large Language Models (LLMs) have significantly enhanced the capabilities of AI systems across a wide range of tasks, thanks to their exceptional ability to process complex semantic and contextual information. The remarkable power of LLMs has inspired exploration into their potential for processing multimodal data, particularly images. Multimodal Large Language Models (MLLMs) typically consist of three main components: a pre-trained large language model, a trained modality encoder, and a modality interface that connects the LLM with encoded modality features.

Recent advances have led to the development of specialized models for mobile and GUI understanding. MobileVLM~\cite{wu_mobilevlm_2024} focuses on enhancing both intra- and inter-UI understanding through targeted pre-training tasks, while MobileFlow~\cite{nong_mobileflow_2024} introduces hybrid visual encoders for variable resolution inputs and multilingual GUI support. These developments have opened new avenues for LLMs in processing GUI tasks across different platforms and languages.

\subsection{GUI Agents and Grounding}

GUI grounding refers to the task of localizing target elements on a screen based on natural language instructions~\cite{cheng_seeclick_2024}. Current approaches primarily fall into two paradigms. The first formulates GUI grounding as a point prediction task, where models directly output coordinates of target elements. Representative works include SeeClick~\cite{cheng_seeclick_2024}, which enhances grounding through pre-training using automated data curation, and ShowUI~\cite{lin_showui_2024}, which introduces UI-guided visual token selection to reduce computational costs.

The second paradigm predicts bounding boxes representing regions that best match instructions, exemplified by works like Aria-UI~\cite{yang_aria-ui_2024}, which adopts a pure-vision approach for GUI grounding. Recent advances have also explored universal visual grounding approaches, such as UGround~\cite{gou_navigating_2025}, which advocates for human-like embodiment where GUI agents perceive environments entirely visually.

Several representative systems have pioneered the GUI agent area. Mobile-Agent~\cite{wang_mobile-agent_2024} explored mobile app usage through autonomous interactions using visual perception, while more recent works have introduced sophisticated frameworks. UI-TARS~\cite{qin_ui-tars_2025} presents a native GUI agent model that solely perceives screenshots and performs human-like interactions, incorporating System-2 reasoning and iterative training with reflective online traces.

\subsection{Reinforcement Learning for GUI Tasks}

Recent advances in GUI agents have shifted toward reinforcement learning-based approaches~\cite{luo_visual_2025, tang_think_2025}. Early works introduced binary hit-or-miss rewards for task completion, while more sophisticated approaches have emerged. UI-R1~\cite{lu_ui-r1_2025} introduces rule-based action rewards and employs Group Relative Policy Optimization (GRPO) for model optimization. GUI-R1~\cite{luo_gui-r1_2025} proposes the first reinforcement learning framework designed to enhance GUI capabilities through unified action space rule modeling.

InfiGUI-R1~\cite{liu_infigui-r1_2025} advances multimodal GUI agents through an Actor2Reasoner framework, transforming agents from reactive actors to deliberative reasoners using spatial reasoning distillation and reinforcement learning. Other notable approaches include ZeroGUI~\cite{yang_zerogui_2025}, which automates online GUI learning at zero human cost, and various methods that employ continuous reward mechanisms for fine-grained feedback~\cite{yuan_enhancing_2025}.

Cross-platform compatibility has emerged as a key challenge in GUI agent development~\cite{xu_mobile-bench-v2_2025}. Works like WebVoyager~\cite{he_webvoyager_2024} focus on web-based interactions, while Mobile-Agent-E~\cite{wang_mobile-agent-e_2025} introduces self-evolving mobile assistants with hierarchical multi-agent frameworks. Recent unified approaches attempt to bridge different domains through shared representations~\cite{huang_spiritsight_2025}.

\subsection{Benchmark Quality and Data Curation}

While substantial progress has been made in developing GUI agents, the quality of evaluation benchmarks has received limited attention~\cite{li_effects_2024}. Most existing work focuses on expanding dataset scale or improving model architectures, assuming that existing benchmarks provide reliable evaluation~\cite{xu_mobile-bench-v2_2025}. However, recent studies have begun to highlight the importance of benchmark quality for fair model evaluation.

The issue of benchmark reliability has started to emerge in the GUI domain. Works like LearnAct~\cite{liu_learnact_2025} focus on demonstration-based learning and highlight the importance of high-quality training data, while A3~\cite{chai_a3_2025} presents evaluation platforms that address limitations in existing benchmarks by providing more realistic task scenarios.

Several recent surveys have provided comprehensive overviews of the field. Tang et al.~\cite{tang_survey_2025} present a systematic examination of (M)LLM-based GUI agents, analyzing their architectural foundations and evaluation methodologies. Liu et al.~\cite{liu_llm-powered_2025} survey LLM-powered GUI agents specifically in phone automation, highlighting the evolution from script-based automation to intelligent, adaptive systems.

Our work addresses this gap by systematically analyzing and improving the quality of GUI agent benchmarks, demonstrating that benchmark deficiencies can significantly impact the assessment of model capabilities. This aligns with broader efforts in the GUI agent community to improve evaluation practices and benchmark reliability.

\section{Conclusion}

In this paper, we addressed a fundamental challenge in GUI agent evaluation: systemic deficiencies in widely-used benchmarks that lead to unfair assessment of model capabilities. Our analysis revealed that biased evaluation metrics and factual errors in data labels collectively result in severe underestimation of existing GUI agents' true potential.

We introduced a rigorous, semi-automated benchmark purification pipeline with two key stages: replacing overly strict point-based matching with intent-aligned bounding box evaluation, and employing consensus failure detection, LLM-based analysis, and human expert verification to correct data-level deficiencies. This process produced AndroidControl-Curated, a high-quality benchmark that provides more accurate reflection of model capabilities.

This finding challenges the dominant paradigm and indicates that future research should prioritize enhancing the quality of benchmarks, rather than focusing solely on benchmark performance metrics. Our work also demonstrates that data quality fundamentally trumps data quantity in training high-performance GUI agents. Our Magma-R1 model, trained on only 2,400 carefully curated samples using our proposed GRPO framework, achieved state-of-the-art performance comparable to models trained on over 31k samples of less rigorously curated data. 

We release AndroidControl-Curated and Magma-R1 as open-source contributions to catalyze a shift toward more rigorous evaluation practices in GUI agent research. Our work demonstrates that addressing fundamental evaluation issues can unlock the true potential of existing models and enable more reliable progress in autonomous GUI interaction.

\begin{acks}
We thank the anonymous reviewers for their valuable feedback and suggestions. This work was made possible by the generous support of several organizations. We extend our sincere gratitude to ArcherMind for providing the high-performance computing resources essential for our experiments. We would also like to acknowledge the BMW Group for their significant administrative support. Furthermore, we are grateful to BA Techworks for invaluable technical support and collaboration throughout this project.
\end{acks}


\bibliographystyle{ACM-Reference-Format}
\bibliography{sample-base}






\end{document}